\PassOptionsToPackage{hidelinks}{hyperref}
\documentclass[journal ]{new-aiaa}
\usepackage[utf8]{inputenc}
\usepackage{textcomp}

\usepackage{multirow}

\usepackage{graphicx}
\usepackage{subcaption}
\usepackage{amsmath}
\usepackage{algorithm}
\usepackage{algpseudocode}

\usepackage{nameref}

\usepackage[version=4]{mhchem}
\usepackage{siunitx}
\usepackage{longtable,tabularx}
\newcommand{\ba}{\emph{\textbf{a}}}
\newcommand{\bb}{\emph{\textbf{b}}}
\newcommand{\bc}{\emph{\textbf{c}}}
\newcommand{\be}{\emph{\textbf{e}}}

\newcommand{\bs}{\emph{\textbf{s}}}
\newcommand{\bx}{\emph{\textbf{x}}}

\newcommand{\bp}{\emph{\textbf{p}}}

\newcommand{\bA}{\emph{\textbf{A}}}
\newcommand{\bI}{\emph{\textbf{I}}}
\newcommand{\bJ}{\emph{\textbf{J}}}

\newcommand{\bS}{\emph{\textbf{S}}}
\newcommand{\bT}{\emph{\textbf{T}}}

\newcommand{\bbeta}{\boldsymbol{\beta}}

\newcommand{\btheta}{\boldsymbol{\theta}}
\newcommand{\bphi}{\boldsymbol{\phi}}
\newcommand{\bpsi}{\boldsymbol{\psi}}
\newcommand{\bomega}{\boldsymbol{\omega}}

\newcommand{\RR}{\mathbb{R}}

\NewDocumentCommand{\COMMENT}{ O{Anonymous} m}{\textcolor{blue}{{[ \sc{#1:~#2} ]}}} 
\NewDocumentCommand{\SUGGEST}{ O{Anonymous} m}{\textcolor{red}{{[#1:~#2]}}} 

\title{Runge-Kutta Approximations for Direct Coning Compensation Applying Lie Theory}

\author{John A. Christian \footnote{Associate Professor, Guggenheim School of Aerospace Engineering, Associate Fellow AIAA.} }
\affil{Georgia Institute of Technology, Atlanta, GA 30332 USA}
\author{Michael R. Walker II\footnote{Principal Member Technical Staff}\footnote{Corresponding author, mrw2@ieee.org}}
\affil{Sandia National Laboratories, Albuquerque, NM 87185 USA}
\author{Wyatt Bridgman\footnote{Now, Senior Data Scientist with C3 AI, Falls Church, VA} } 
\affil{Sandia National Laboratories, Livermore, CA 94551 USA}
\author{Michael J. Sparapany\footnote{Principal Member Technical Staff}}
\affil{Sandia National Laboratories, Albuquerque, NM 87185 USA}

\begin{document}

\maketitle



\section{Introduction}
Strapdown inertial measurement units (IMUs) are ubiquitous sensors in modern aerospace systems, automobiles, ground robotics, consumer electronics, and other platforms. Typical IMUs are composite sensors consisting of both gyroscopes (gyros) and accelerometers. In their simplest configuration, a three-axis IMU consists of three orthogonally arranged gyros and three orthogonally arranged accelerometers. Each gyro within the IMU measures the instrument’s inertial angular velocity in a particular instrument-fixed direction. Likewise, each accelerometer measures the instrument’s sensed acceleration (specific force). This work focuses primarily on the integration of the gyro measurements to obtain an estimate of the inertial change in IMU orientation. Such IMU integration is a central component to the propagate function within most aerospace navigation filters \cite{Markley:2014,crassidisThreeAxisAttitudeEstimation2016a, carpenterNavigationFilterBest2025, christianFundamentalsSpacecraftOptical2026}.

\begin{figure}[t]
\centering
\includegraphics[width=0.75\textwidth]{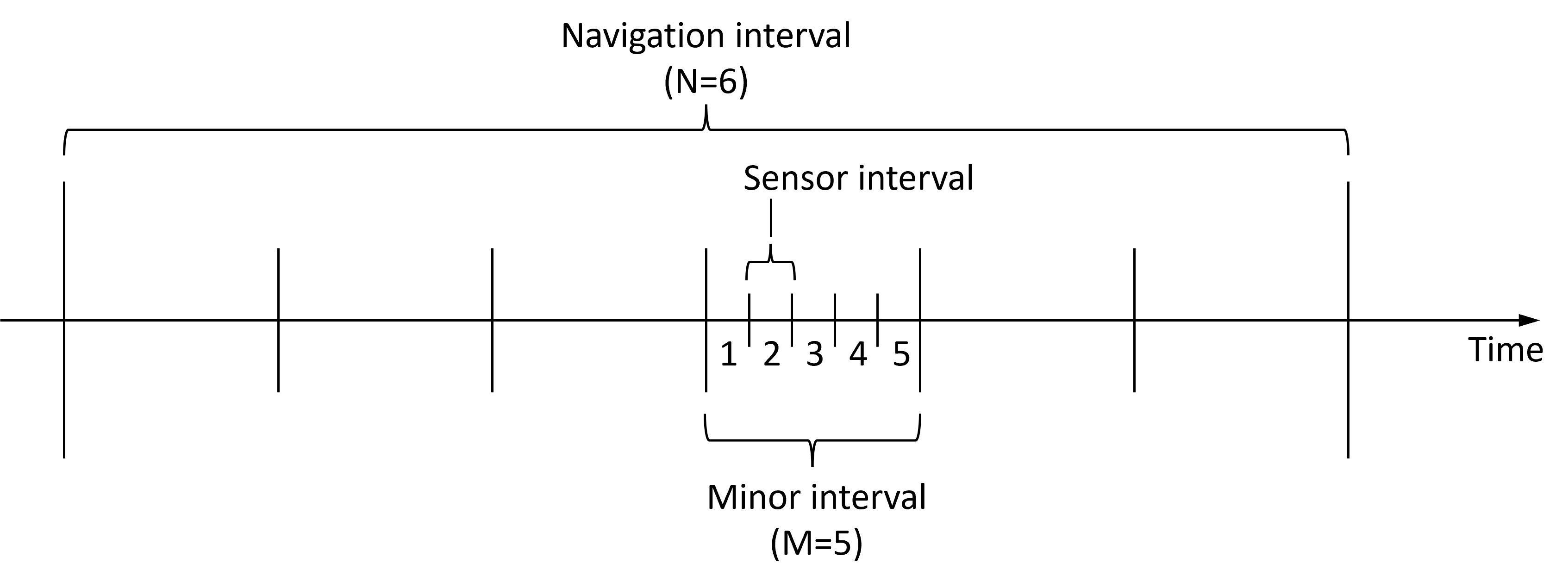}
    \caption{Illustration of time intervals.}
    \label{fig:TimeIntervales}
\end{figure}

The conventional approach for integrating gyro measurements follows a multi-frequency scheme. This scheme, which has been written about extensively elsewhere \cite{Ignagni:1990,Savage:1998}, consists of three important time intervals (Fig.~\ref{fig:TimeIntervales}). The shortest interval is the sensor interval (sometimes called the ``subminor interval''), corresponding to the time over which the instrument integrates the raw gyro signal to produce a discrete measurement for processing. These discrete measurements are produced at a high rate, generally on the order of several kHz. The middle interval is the minor interval in which the kHz measurements are integrated to produce a rotation measurement suitable for consumption by a navigation filter. This is usually done within the IMU and the minor interval corresponds to the time between measurements of the IMU data stream seen by a user. Finally, the navigation interval (sometimes called the ``major interval'') is the time between state estimates within the navigation filter.

The raw signal from each gyro is typically integrated in hardware on a per-channel basis. The independent integration of gyro channels introduces error in a rotating strapdown IMU since the gyros change orientation over the course of integration and rotations are noncommutative. Error introduced by this process is referred to as \emph{coning error} and corrections are attempted through coning compensation algorithms\cite{Ignagni:1996, Savage:1998}. Many older IMUs also integrate gyro measurements on a per-channel basis over the minor interval due to computational constraints in keeping up with a kHz data stream and then perform higher-order coning corrections using a multitude of sensor-interval integrations \cite{Savage:1998,Ignagni:1996}. Although many IMUs continue to separate the commutative and noncommutative integrations at the minor interval, this no longer necessary with modern computational capabilities. Nevertheless, coning correction remains necessary over the sensor interval whenever raw gyro signals are integrated (or averaged) on a per-channel basis.

The main contribution of this work is generalizing coning compensation algorithms to higher-order integrators.  Gyros do not provide absolute measures of vehicle attitude. Instead, vehicle attitude states must be propagated - or integrated - using inertial measurements. Attitude propagation fundamentally represents an initial value problem where the ordinary differential equation (ODE) depends on the coordinate choice for state variables. Quaternions are a common choice for attitude representation within the navigation community, and high-order integration methods have been demonstrated for instantaneous rate measurements \cite{andrleGeometricIntegrationQuaternions2013a}. Low-order methods have been demonstrated for rate-integrating gyro \cite{crassidisThreeAxisAttitudeEstimation2016a}. For attitude represented in exponential coordinates (also known as the rotation vector), the ODE is referred to as the Bortz equation in the navigation community. This work makes explicit the equivalence between the Bortz equation and the inverse-Jacobian of the Lie algebra. Recognizing the Bortz equation as an invertible matrix simplifies the state-propagation ODE. The simplified ODE conveniently fits into traditional Runge-Kutta integrators, and approximate solutions are available by discarding higher-order terms. The Lie-theoretic perspective highlights a need to model angular rates from integrated measurements. The second contribution is fitting polynomial models for instantaneous rates from integrated measurements. Aggregating multiple measurements enables higher-order polynomial models which in-turn enables higher-order integrator methods. The third contribution is quantifying variations in integration error based on solver order, polynomial order of rate models, and approximate solutions to the ODE. New methods for propagating attitude are provided for inertial navigation leveraging consecutive integrated rate measurements. The combination of curve-fitting and ODE solver selection decouples solver order from the number of integrated measurements. Results can be used to reduce integration error or as a method to increase step size (potentially reducing computations) while preserving fidelity.

\section{Attitude Representations and Kinematics}

Consider an IMU with body frame $B$ that is rotating with respect to an inertial frame $I$ at an angular velocity of $\bomega$. This notation assumes that $\bomega$ is expressed in the rotating frame $B$. The rotation between frames $I$ and $B$ may be written using the direction cosine matrix (DCM) $\bT$, such that a vector $\bx$ is transformed according to 
\begin{equation}
	\label{eq:Usage}
	\bx_B = \bT \bx_I
        \quad \text{and} \quad
        \bx_I = \bS \bx_B
\end{equation}
where $\bx_I$ is the vector expressed in frame $I$ and $\bx_B$ is the vector expressed in frame $B$. Here, the passive interpretation of attitude is assumed such that $\bT$ describes a change of basis \cite{Markley:2014,Zanetti:2019} transforming a vector expressed in frame $I$ to that same vector expressed in frame $B$. The DCM $\bS = \bT^{-1}$ describes the reverse transformation.


Every DCM $\bT$ (or $\bS$) belongs to the group $SO(3)$ consisting of $3 \times 3$ proper orthogonal matrices. As a consequence, every DCM $\bT \in SO(3)$  has three free parameters and six constraints. Although the DCM is convenient for performing frame transformations of vectors, the constraints make perfectly on-manifold integration of the DCM more difficult. It is preferable to perform the integration on a minimal attitude representation (i.e., a three-parameter representation). Many such representations exist \cite{Markley:2014}, but only the rotation vector is considered here for brevity.

The problem of attitude representation, kinematics, and integration may be interpreted through either a conventional aerospace framework or from a Lie-theoretic framework. These interpretations are equivalent and lead to exactly the same equations. However, as will be shown, each also comes with its own history, vocabulary, and observations.

\subsection{Conventional Aerospace Framework}
Euler's rotation theorem states that any arbitrary 3-axis rotation may be described by the rotation about a single axis $\be$ and a single angle $\varphi$ \cite{eulerFormulaeGeneralesPro1776, Markley:2014, christianFundamentalsSpacecraftOptical2026}. Taking the product of these forms the rotation vector $\bphi$, which is a $3 \times 1$ vector pointing in the direction of $\be$ with a magnitude of $\varphi$. Thus, the rotation vector is written as
\begin{equation}
    \label{eq:RotVecDef}
	\bphi = \varphi \be
\end{equation}
with kinematics of
\begin{equation}
    \label{eq:BortzEqn}
    \dot{\bphi} = \bomega + \frac{1}{2} \bphi \times \bomega + \frac{1}{\varphi^2}\left( 1 - \frac{\varphi \sin \varphi}{2(1-\cos \varphi)} \right)\left[ \bphi \times (\bphi \times \bomega) \right]
\end{equation}
where $\bomega$ is the angular velocity of frame $B$ with respect to frame $I$ as expressed in frame $B$. Since $\bphi$ is a minimal parameterization, singularities exist. However, for the exponential coordinates singularities are not encountered for ${\varphi < 2\pi}$ \cite{stuelpnagelParametrizationThreeDimensionalRotation1964}. 

Equation~\eqref{eq:BortzEqn} is sometimes referred to as the ``Bortz equation'' (especially in the IMU literature) after an influential derivation presented by the American engineer John Bortz in 1971 \cite{Bortz:1971}. A somewhat simpler way of arriving at the same result using quaternions may be found in Ref.~\cite{Markley:2014}.

The elements $\be$ and $\varphi$ of the rotation vector are related to the DCM through the finite rotation formula \cite{Markley:2014}
\begin{equation}
    \label{eq:EulerFiniteRotation}
	\bT(\be,\varphi) = \bI_{3 \times 3} - (\sin \varphi) [\be \times] + (1 - \cos \varphi) [\be \times ]^2
\end{equation}
where $[\, \cdot \, \times]$ is the right-handed, skew-symmetric, cross product matrix satisfying $\ba \times \bb = [\ba \times] \bb$ with elements
\begin{equation}
	[\bx \times] = 
	\begin{bmatrix}
		0 & -x_3 & x_2 \\
                  x_3 & 0 & -x_1 \\
                  -x_2 & x_1 & 0
	\end{bmatrix}
\end{equation}
 The DCM may also be written as the matrix exponential
\begin{subequations}
\label{eq:MatExpDCMboth}
\begin{equation}
    \label{eq:MatExpDCM}
	\bT(\bphi) = \exp\left( [-\bphi \times ]  \right)
\end{equation}
\begin{equation}
	\bS(\bphi) = \exp\left( [\bphi \times ]  \right)
\end{equation}
\end{subequations}
though most navigators compute the DCM in practice using eq.~\eqref{eq:EulerFiniteRotation} for computational reasons. The DCM kinematics are also recognized to be
\begin{subequations}
    \begin{equation}
    \label{eq:KinematicsDCM}
    \dot{\bT} = \left[ -\bomega \times \right] \bT
\end{equation}
\begin{equation}
    \label{eq:KinematicsDCMS}
    \dot{\bS} =  \bS \left[ \bomega \times \right]
\end{equation}
\end{subequations}
such that $\dot{\bT}^\top = \dot{\bS}$. All of these notions are routinely encountered in aerospace textbooks \cite{Markley:2014,Crassidis:2012}.
The reader is reminded that $\bomega \neq \dot{\bphi}$, in general, with an important exception of $\bphi = \textbf{0}_{3 \times 1}$. These things will be important later.

\subsection{Lie-Theoretic Framework}
Each of the key equations for attitude representation and kinematics appearing in the conventional aerospace framework above may also be produced using Lie algebra. There should be no surprise that the different framework produces the same equations, as this must be the case! The Lie-theoretic interpretation will become helpful in developing higher-order integrators and coning corrections. Thus, let us briefly rewrite attitude representations and their kinematics in the language and notation of Lie algebra. 

One of the primary advantages of the Lie-theoretic approach is that an exponential map can be used to relate a Lie algebra with a corresponding Lie group. Although extensions to other Lie groups are possible, this work focuses strictly on $SO(3)$. The corresponding Lie algebra $so(3)$ may be constructed with the wedge operator $( \, \cdot \, )^\wedge : \mathbb{R}^3\rightarrow so(3)$. The exponential map $\exp : so(3) \rightarrow SO(3)$ relating $\bT \in SO(3)$ and $\bs^{\wedge} \in so(3)$ is simply the matrix exponential
\begin{equation}
    \label{eq:MatExpLie}
    \bT = \exp( \bs^{\wedge} )
\end{equation}
where $\bs$ is commonly referred to as ``exponential coordinates'' in the Lie algebra literature. Here, the exponential coordinates relate to the rotation vector from eq.~\eqref{eq:RotVecDef} as ${\bs = -\bphi}$. Thus, if $\bphi$ is an angle vector describing a change of basis from frame $I$ to frame $B$, then it follows that $\bs$ is an angle vector describing a change of basis from frame $B$ to frame $I$. It is evident from comparison with eq.~\eqref{eq:MatExpDCMboth} that the wedge operator is simply a skew-symmetric matrix given by
\begin{equation}
\bs^\wedge = \left[\bs \times \right]
\end{equation}
when speaking of $so(3)$. This is well-known \cite{Wang:2008,Forster:2017}. It is worth emphasizing, however, that group transformations extend elegantly to higher-dimensional spaces and other groups with relevance to navigation and robotics. For example, unit-quaternions compose a Lie Group  \cite{andrleGeometricIntegrationQuaternions2013a}. The groups $SO(4)$ and $SE(3)$ are both inherently 6-dimensional with corresponding Lie Algebras $so(4)$ and $se(3)$, respectively. See Ref.~\cite{chirikjian_stochastic_2012} for a concise generalized treatment.

The kinematics from eq.~\eqref{eq:KinematicsDCM} can be restated for $\bS$ using Lie-theoretic notation \cite{iserles2000lie,chirikjian_stochastic_2012}
\begin{equation}
    \label{eq:KinematicsLie}
    \dot{\bT} = -\bomega^\wedge \,\bT 
\end{equation}
Since $\bT^{-1} = \bT^\top \in SO(3)$, one implication of eq.~\eqref{eq:KinematicsLie} is that $\bT \dot{\bT}^\top = \bomega^{\wedge} \in so(3)$. As a reciprocal operation to $( \, \cdot \, )^\wedge$, the ``vee operator'' converts skew-symmetric matrices to their vector representations ${( \, \cdot \, )^\vee : so(3) \rightarrow \mathbb{R}^3}$. Now $\bT \dot{\bT}^\top$ can be expressed as a vector decomposed as a matrix-vector product
\begin{equation}
\left(\bT \dot{\bT}^\top  \right)^\vee = \left(\bomega^\wedge\right)^\vee = \bomega
\end{equation}

This framework may be used to relate the angle vector rates $\dot{\bphi}$ or $\dot{\bs}$ with the angular velocity $\bomega$. Much of the Lie algebra literature focuses on $\bs$, but it is more convenient write things in terms of $\bphi$ here. This is achieved by introducing the right-Jacobian $\bJ_{\bphi}$ (sometimes referred to as $\text{d}\exp$), such that
\begin{equation}
    \label{eq:ode_omega}
	\bomega = \bJ_{\bphi} \dot{\bphi}
\end{equation}
Since the Jacobian is invertible for all $\bs$ \cite{chirikjian_stochastic_2012},  the differential equation in the Lie algebra is simply
\begin{equation}
\label{eq:ode_thetadot}
\dot{\bphi} = \bJ_{\bphi}^{-1} \bomega
\end{equation}
For $SO(3)$, the inverse right-Jacobian $\bJ_{\bphi}^{-1}$ has the closed form (c.f. equation (10.86) \cite{chirikjian_stochastic_2012}, also \cite{hausdorff1906symbolische,iserles1984solving, Muller:2023,sola2018micro})\footnote{Note that the sign of the middle term $(1/2)\bphi^\wedge$ may vary from one reference to another depending on the choice of an active or passive representation of attitude.   }
\begin{align}
\label{eq:Jinv}
    \bJ_{\bphi}^{-1} =  \bI + \frac{1}{2}\bphi^\wedge +\left( \frac{1}{\varphi^2} - \frac{1+\cos \varphi}{2 \varphi \sin \varphi}\right)\left(\bphi^\wedge\right)^2 
\end{align}
It is helpful to observe \cite{Savage:1998}
\begin{equation}
	\label{eq:Trig_equality}
    \left( \frac{1}{\varphi^2} - \frac{1+\cos \varphi}{2s \sin \varphi}\right) = 
    \left( \frac{1}{\varphi^2} - \frac{\sin \varphi}{2 \varphi (1 - \cos \varphi)}\right)
    = \frac{1}{\varphi^2}\left( 1 - \frac{\varphi \sin \varphi}{2(1-\cos \varphi)} \right)
    \approx
    \frac{1}{12}\left( 1 + \frac{1}{60}\varphi^2 + \hdots \right)
\end{equation}
such that to third order in $\varphi = \| \bphi \|$ the right-Jacobian may be approximated as
\begin{equation}
    \bJ_{\bphi}^{-1} \approx \bI + \frac{1}{2}\bphi^\wedge + \frac{1}{12}\left(\bphi^\wedge\right)^2
    \label{eq:Jinv:approx}
\end{equation}

It is now a straightforward task to show the equivalence of the Lie-theoretic result to the Bortz equation. Substituting eq.~\eqref{eq:Jinv} and eq.~\eqref{eq:Trig_equality} into eq.~\eqref{eq:ode_thetadot},
\begin{equation}
	\label{eq:ode_thetadot_expanded}
	\dot{\bphi} = \bJ^{-1}_{\bphi} \bomega = \left[ \bI + \frac{1}{2}\bphi^\wedge +\frac{1}{\varphi^2}\left( 1 - \frac{\varphi \sin \varphi}{2(1-\cos \varphi)} \right)\left(\bphi^\wedge\right)^2\right] \bomega
\end{equation}
which is equivalent to \eqref{eq:BortzEqn}. Recalling that $\bphi^{\wedge} = [\bphi \times]$, and using ${\bphi \times \left[\bphi \times \bomega\right] = \left(\bphi^\wedge \right)^2\bomega}$, produces the desired equivalence.

For completeness,
\begin{equation}
	J_{\bphi}  =   \bI - \frac{ 1 - \cos \varphi}{\varphi^2}\bphi^\wedge + \frac{\varphi - \sin \varphi}{\varphi^3} \left(\bphi^\wedge \right) ^2
\end{equation}
using the right-Jacobian convention \cite{chirikjian_stochastic_2012}.

The limit of \eqref{eq:Trig_equality} as ${\varphi\rightarrow 0}$ is simply $1/12$. Consequently, the Jacobian is well behaved over the domain  ${\bphi < 2\pi}$ when $\bphi$ is expressed in exponential coordinates. For other three-dimensional parameterizations of rotations (e.g., Euler angles), singularity are encountered in similar domains \cite{stuelpnagelParametrizationThreeDimensionalRotation1964}.

\section{Conventional Coning Correction}
Rather than directly integrating the kinematics, most real-time systems use IMUs to obtain an estimate of an incremental rotation vector $\Delta \bphi_k = \bphi_k - \bphi_{k-1}$ over the corresponding time interval $\Delta t_k = t_k - t_{k-1}$. Then, the attitude is advanced from one time to the next according to
\begin{equation}
	\bT_k =\bT( \Delta \bphi_k) \, \bT_{k-1}
\end{equation}
Computation of $\bT( \Delta \bphi_k)$ may use either the finite rotation formula from eq.~\eqref{eq:EulerFiniteRotation} or the exponential map from eqs.~\eqref{eq:MatExpDCM} or \eqref{eq:MatExpLie}. In practice the former is arguably more prevalent, though it does not matter much one way or the other. 

\begin{figure}[h!]
\centering
\includegraphics{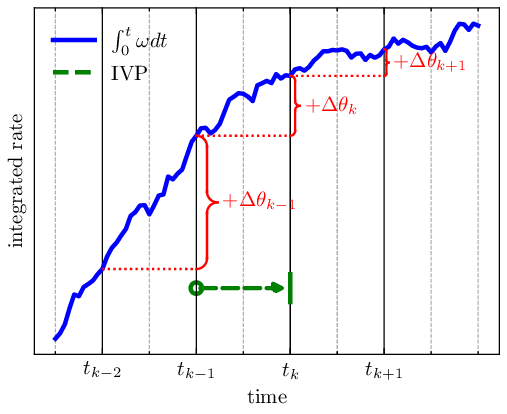}
\caption{Sampling of integrated rate and integration period of interest.}
\label{fig:sampling}
\end{figure}

Ideally, the value of $\Delta \bphi_k = \Delta \varphi_k \, \be$ would be found by integrating $\dot{\bphi}$ from eqs.~\eqref{eq:BortzEqn} or \eqref{eq:ode_thetadot} with an initial condition of $\bphi = \textbf{0}_{3 \times 1}$ such that, for every time step, 
\begin{equation}
        \label{eq:DefDeltaPhi}
	\Delta \bphi_k = \int_{t_{k-1}}^{t_k} \dot{\bphi} \, dt
\end{equation}
What makes this challenging in practice is that the gyros within an IMU do not directly measure either $\Delta \bphi_k$ or $\dot{\bphi}$. Instead, in an ideal IMU with perfectly orthogonal gyros, each gyro channel measures $\omega_1$, $\omega_2$, or $\omega_3$ in the rotating body frame. The measurements compose the angular velocity vector $\bomega = [\omega_1; \omega_2; \omega_3]$. Thus, per-channel integration of the angular velocity produced by a typical rate-integrating gyro (e.g., a ring laser gyro) actually produces a measurement of
\begin{equation}
	\label{eq:PhiIntegralDef}
	\btheta(t) = \int_{t_{0}}^t \bomega \, d \tau
\end{equation}
which, over any specific interval of time, is
\begin{equation}
	\label{eq:DefDeltaTheta}
	\Delta \btheta_k = \int_{t_{k-1}}^{t_k }\bomega \, d \tau
\end{equation}
Importantly, since $\dot{\bphi} \neq \bomega$, it follows that $\Delta \bphi_k \neq \Delta \btheta_k$. The relationship between integrated $\omega$ and $\Delta\theta$ are illustrated in Figure \ref{fig:sampling}.

The objective is to write the desired quantity ($\Delta \bphi_k$) in terms of available measurements ($\Delta \btheta_k$). The difference in the two is principally due to the noncommutitavity terms in the Bortz equation [the right-most term in eq.~\eqref{eq:BortzEqn}]. Thus, integrating the Bortz equation to obtain $\bphi$ produces
\begin{equation}
 \bphi = \underbrace{\int_{t_{0}}^{t} \bomega \, d\tau}_{\btheta} +\underbrace{\int_{t_{0}}^{t} \left\{ \frac{1}{2} \bphi \times \bomega + \frac{1}{\varphi^2}\left( 1 - \frac{\varphi \sin \varphi}{2(1-\cos \varphi)} \right) \left[ \bphi \times (\bphi \times \bomega) \right] \right\} \, d\tau}_{\bbeta}
\end{equation}
or, simply,
\begin{equation}
    \label{eq:BortzSuperSimple}
    \bphi = \btheta + \bbeta
\end{equation}
The \emph{coning correction} term $\bbeta$ is defined as
\begin{equation}
    \label{eq:ConingFull}
 \bbeta =\int_{t_{0}}^{t} \left\{ \frac{1}{2} \bphi \times \bomega + \frac{1}{\varphi^2}\left( 1 - \frac{\varphi \sin \varphi}{2(1-\cos \varphi)} \right) \left[ \bphi \times (\bphi \times \bomega) \right] \right\} \, d\tau
\end{equation} 
Thus, as is clear from eq.~\eqref{eq:BortzSuperSimple}, the coning term $\bbeta$ is a compensation to the measured $\btheta$ that attempts to account for the changing orientation of frame $B$ over the integration. Clearly, for a one dimensional problem where $\bomega$ remains fixed in the body frame, it is found that $\bbeta_k = \textbf{0}$ since $\bphi \times \bomega = \textbf{0}$.
 
\subsection{Simplification of the Coning Correction}
When the rotation vector is small (which it often is for small time steps since integration is always started from $\bphi = \textbf{0}$), the coefficient of the right-hand term may be expanded as a Taylor series in the same was as eq.~\eqref{eq:Trig_equality} to find
\begin{equation}
	 \frac{1}{\varphi^2}\left( 1 - \frac{\varphi \sin \varphi}{2(1-\cos \varphi)} \right) = \frac{1}{12} \left( 1 + \frac{1}{60} \varphi^2 + \hdots \right)
\end{equation}
Thus, to third order in $\varphi$, the coning correction simplifies to
\begin{equation}
    \label{eq:ConingMiddle}
 \bbeta \approx \int_{t_{0}}^{t} \left\{ \frac{1}{2} \bphi \times \bomega + \frac{1}{12} \left[ \bphi \times (\bphi \times \bomega) \right] \right\} \, d\tau
\end{equation}
which is an excellent approximation in most cases. More than this, the so-called Goodman-Robinson approximation \cite{Goodman:1958} maintains only first-order terms in the attitude and replaces $\bphi$ with $\btheta$ inside the integrand
\begin{equation}
    \label{eq:ConingGR}
 \bbeta \approx \frac{1}{2} \int_{t_{0}}^{t}   \btheta \times \bomega \, d\tau
\end{equation}
The replacement of $\bphi$ by $\btheta$ may be viewed as a single iteration within a Picard iteration scheme. The Goodman-Robinson approximation, which predates the Bortz's work by over a decade, is the foundation for many of the classical coning algorithms \cite{Ignagni:1990,Ignagni:1996,Savage:1998}. Some newer algorithms attempt to solve either eq.~\eqref{eq:ConingFull} or eq.~\eqref{eq:ConingMiddle}, though this comes with an increase in complexity \cite{Savage:2006,Wang:2019,Ignagni:2020}.

\subsection{Approximate Solutions to the Coning Correction Term}
Consider a rate integrating gyro producing measurements of $\Delta \btheta_k$ at the sensor interval from eq.~\eqref{eq:DefDeltaTheta}. Suppose that $m$ such measurements $\{ \Delta \btheta_k \}_{k=1}^m$ are accumulated over an IMU's minor interval. The objective is to integrate this set of IMU observations to produce an estimate of the DCM $\bT_m$ or the rotation vector $\bphi_m$ at the end of the minor interval. The Goodman-Robinson approximation from eq.~\eqref{eq:ConingGR} may be used to solve this problem analytically if some assumptions on the time-history of $\bomega$ are made. The simplest non-trivial assumption is a linearly varying angular velocity. 

Most of the classical coning literature focuses on the two-speed paradigm where $m > 1$, where the ``two speeds'' being the rate of sensor interval and the rate of the minor interval (c.f. Fig. \ref{fig:TimeIntervales}). Under the two-speed paradigm, where $m > 1$ sensor interval measurements are obtained over a minor interval, a linear $\bomega$ may be assumed to obtain the classic approximation \cite{Ignagni:1990}
\begin{equation}
    \label{eq:ConingTwoSpeed}
    \bphi_m = \btheta_m + \frac{1}{2}\sum_{k=1}^m \btheta_k \times \Delta \btheta_k + \frac{1}{12} \sum_{k=1}^m \Delta \btheta_{k-1} \times \Delta \btheta_{k}
\end{equation}
where
\begin{equation}
    \btheta_m = \sum_{k=1}^m \Delta \btheta_k
\end{equation}
The two-speed techniques exemplified by eq.~\eqref{eq:ConingTwoSpeed}---though there are many others under different assumptions for the time-history of $\bomega$ \cite{Ignagni:1996,Savage:1998}---are exceptionally fast and remove the need for performing the computationally expensive task of finite rotations at the sensor interval. However, there have been considerable advancements in computing since 1990, and two-speed algorithms are not always necessary. Propagating at the sensor interval is well-known to be desirable when it is possible \cite{Savage:1998}. 

If finite rotations are to be performed at the sensor interval, then coning correction is only applied over the sensor interval (and not the minor interval). This leads to the one-speed algorithm, which may be easily derived for linear $\bomega$ as described in the Appendix and results in
\begin{equation}
    \label{eq:ConingOneSpeed}
    \Delta \bphi_k = \Delta \btheta_k + \frac{1}{12} \Delta \btheta_{k-1} \times \Delta \btheta_{k}
\end{equation}
which is the same correction suggested by Miller in the early 1980s \cite{Miller:1983}. Thus, letting $\Delta \bT_k = \bT( \Delta \bphi_k)$ be computed from eq.~\eqref{eq:EulerFiniteRotation}, the attitude at the end of the sensor interval is then computed as the product
\begin{equation}
	\label{eq:T:expanded}
    \bT_m = \prod_{k=1}^m \Delta \bT_k
    = \Delta \bT_m \cdots \Delta \bT_k \cdots \Delta \bT_1
\end{equation}
and where the product operator is applied right to left.


\section{Numerical Methods for Coning Correction}
\label{sec:ode}
In light of eq.~\eqref{eq:DefDeltaPhi}, coning compensation amounts to solving the differential equation of eq.~\eqref{eq:ode_thetadot} for $\bphi\left(t_k + \Delta t\right)$ from an initial condition $\bphi\left(t_k\right) = 0$. This generalized perspective enables usage of traditional numerical solvers such as Runge-Kutta methods. Indeed, using traditional Runge-Kutta style solvers an identical expression is obatined for the single-speed algorithm fom eq.~\eqref{eq:ConingOneSpeed}. There are three benefits to exploring an alternative derivation. First, equivalence with traditional coning correction methods demonstrates applicability of Runge-Kutta style solvers. Second, generalized solvers accommodate non-zero initial conditions.  Solving for $\Delta \bphi_k$ only requires solving the Bortz equation from an initial condition of all zeros. A generalized approach facilitates solving for  $\bphi_k$ from $\bphi_{k-1}$ directly potentially reducing computations. Thirdly, the recursive structure of Runge-Kutta methods easily generalizes to higher-order algorithms.

\subsection{Runge-Kutta Methods and Sampling Requirements}
The Runge-Kutta algorithm, stated in Algorithm \ref{alg:RK}, is a generalized numerical solver for initial value problems. The generalized ordinary differential equation $f(t,\bphi) : \mathbb{R} \times \mathbb{R}^m \rightarrow \mathbb{R}^m$, is solved from an initial condition ${\bphi}_k\in\mathbb{R}^3$ at time $t_k$ to the final value ${\bphi}_{k+1} = {\bphi}\left(t_{k} + \Delta t\right)$ at stepsize $\Delta t$. 

\begin{algorithm}[H]
	\caption{Runge-Kutta}
	\label{alg:RK}
	\begin{algorithmic}[1]
		\Require $f : \mathbb{R} \times \mathbb{R}^m \rightarrow \mathbb{R}^m$ ordinary differential equation; ${\bphi}_k\in\mathbb{R}^m$ initial condition; $t_k,\Delta t \in\mathbb{R}$: timing parameters; $\bA\in\mathbb{R}^{n \times n}$, $\bb,\bc\in\mathbb{R}^{n}$ Butcher-tableau.
		\State $n \gets$ order of RK scheme
		\For{$\nu=1$ to $n$}
		\State $\bpsi_\nu \gets {\bphi}_k + \sum_{l=1}^{\nu-1} \bA_{\nu,l}{\bf f}_l$
		\State ${\bf f}_\nu \gets \Delta t f(t_k + \Delta t \bc_\nu, \bpsi_\nu)$ \label{alg:Rk:sample}
		\EndFor
		\State $\bphi_{k+1} \gets \bphi_k + \sum_{l=1}^n \bb_l {\bf f}_l$
	\end{algorithmic}
\end{algorithm}

Algorithm \ref{alg:RK} extends to different solver orders and sampling schemas as determined by $\bA\in\mathbb{R}^{n \times n}$, $\bb,\bc\in\mathbb{R}^{n}$. Butcher tableaux \cite{butcherRungeKuttaProcessesHigh1964,butcherNumericalMethodsOrdinary2016} are useful for depicting the algorithm hyperparameters $\bA$,$\bb$,$\bc$ with examples shown in Figs.~\ref{tbl:fwdeuler} - \ref{tbl:rk4} for common implementations.  Note, the array $\bc$, depicted in tableau's first column, identifies both the solver order and the timesteps at which point the differential equation, $f$, is evaluated. The sample points are listed relative to step size $\Delta t$ (c.f. ${t_0 + \Delta t \bc_\nu}$ in Algorithm \ref{alg:RK}, line \ref{alg:Rk:sample}).

\begin{figure}[h]
	\centering
	\begin{tabular}
		{c|c}
		0 & 0\\
		\hline
		& 1 
	\end{tabular}
 \caption{Forward Euler Tableau}
	\label{tbl:fwdeuler}
\end{figure}

\begin{figure}[h]
	\centering
	\begin{tabular}
		{c|cc}
		0 & 0 & 0\\
		1/2 & 1/2 & 0 \\
		\hline
		& 0 & 1
	\end{tabular}
    \caption{Explicit-Midpoint Tableau}
	\label{tbl:expmid}
\end{figure}

\begin{figure}[h]
	\centering
	\begin{tabular}
		{c|ccc}
		0 & 0 & 0 & 0 \\
		1/2 & 1/2 & 0  & 0\\
		1 & -1 & 2 & 0\\
		\hline
		& 1/6 & 2/3 & 1/6
	\end{tabular}
 \caption{Runge-Kutta Original Third Order Tableau}
	\label{tbl:rk3}
\end{figure}

\begin{figure}[h]
	\centering
	\begin{tabular}
		{c|cccc}
		0 & 0 & 0 & 0 & 0\\
		1/2 & 1/2 & 0  & 0 & 0\\
		1/2 & 0 & 1/2 & 0  & 0\\
		1 & 0 & 0 & 1 & 0\\
		\hline
		& 1/6 & 1/3 & 1/3 & 1/6
	\end{tabular}
    \caption{Runge-Kutta Original Fourth Order Tableau}
	\label{tbl:rk4}
\end{figure}

To solve eq.~\eqref{eq:DefDeltaPhi}, using Algorithm \ref{alg:RK}, eq.~\eqref{eq:ode_thetadot} is employed as
\begin{equation}
	\label{eq:f}
	f(t,\bphi ) \coloneq \bJ^{-1}_{\bphi} \bomega (t)
\end{equation}
In this case, the $f$ separates as two terms. The first term, $\bJ^{-1}_{\bphi}$, depends on the dependent variable $\bphi$. The second term $\bomega(t)$, for inertial navigation, is obtained through sensor measurements. Notice, Algorithm \ref{alg:RK} samples $\bomega(t)$ at discrete points as determined by $t_k$, $\Delta t$, and $\bc$. First Runge-Kutta methods are expanded assuming direct access to $\bomega$ for any $t$.

Applying the {\it forward-Euler} method, calling Algorithm \ref{alg:RK} using Fig.~\ref{tbl:fwdeuler}, yields
\begin{equation}
	\hat{\bphi}_\textrm{FwdE}\left(t_k+\Delta t\right) = \bphi_k + \Delta t \, \bJ_{\bphi_k}^{-1}\bomega\left(t_k\right)
\end{equation}
The {\it explicit midpoint} method makes use of both $\bomega\left(t_k\right)$ and ${\bomega\left(t_k + \frac{\Delta t}{2}\right)}$
\begin{equation}
	\label{eq:bphi:expmid}
	\hat{\bphi}_\textrm{ExMid}\left(t_k + \Delta t\right) = \bphi_k + \Delta t \, \bJ_{\bpsi }^{-1}\bomega\left(t_k + \frac{\Delta t}{2}\right)
\end{equation}
where the inverse-Jacobian requires an intermediate
\begin{equation}
	\bpsi = \bphi_k + \frac{\Delta t}{2} \bJ_{\bphi_k}^{-1}\bomega\left(t_k\right)
\end{equation}
Motivated by eq.~\eqref{eq:DefDeltaPhi}, $\bphi_k = 0$ is selected which effects ${\bJ_{\bphi_k}^{-1}=\bI_{3 \times 3}}$. Consequently, 
\begin{align}
	\Delta\hat{\bphi}_\textrm{FwdE}\left(t_k+\Delta t\right) &= \Delta t \, \bomega\left(t_k\right)\\
	\Delta\hat{\bphi}_\textrm{ExMid}\left(t_k + \Delta t\right) &= \Delta t \, \bJ_{\Delta \bpsi}^{-1} \, \bomega\left(t_k + \frac{\Delta t}{2}\right)\\
	\Delta\bpsi &= \frac{\Delta t}{2} \bomega\left(t_k\right)
\end{align}
For higher-order methods, expanding Algorithm \eqref{alg:RK} becomes tedious. However, using the initial condition $\bphi(t_k)={\bf 0}$, the small-angle approximation eq.~\eqref{eq:Jinv:approx}, and retaining only quadratic terms,
\begin{align}
	\label{eq:rk3:omega}
	\Delta\hat{\bphi}_\textrm{RK3} \approx & \frac{\Delta t}{6} \left( \bomega\left(t_k\right) + 4\bomega\left(t_k + \frac{\Delta t}{2}\right) + \bomega\left(t_k+\Delta t\right) \right) \nonumber \\
	& + \frac{\Delta t^2}{6}\left(\bomega\left(t_k\right) - \bomega\left(t_k+\Delta t\right)\right)\times \bomega\left(t_k+ \frac{\Delta t}{2}\right) + \frac{\Delta t^2}{12}\bomega\left(t_k+\Delta t\right)\times \bomega\left(t_k+ \frac{\Delta t}{2}\right)
\end{align}
and
\begin{align}
	\label{eq:rk4:omega}
	\Delta\hat{\bphi}_\textrm{RK4} \approx \frac{\Delta t}{6} \left( \bomega\left(t_k\right) + 4\bomega\left(t_k + \frac{\Delta t}{2}\right) + \bomega\left(t_k+\Delta t\right) \right)
	 + \frac{\Delta t^2}{12}\left(\bomega\left(t_k\right) - \bomega\left(t_k+\Delta t\right)\right)\times \bomega\left(t_k+ \frac{\Delta t}{2}\right)
\end{align}
are obtained. Samples of $\bomega$ must be provided either through direct measurements (e.g., gyros providing instantaneous rates) or from a model (e.g., parameterized from integrated rate measurements).


\subsection{Approximating $\bomega$ from $\Delta\btheta$}
\label{sec:ApproximatingOmega}
When $\bomega$ is available, eq.~\eqref{eq:rk4:omega} may be used directly. However, for rate-integrating gyros, $\bomega$ must be modeled from $\Delta\btheta$ measurements. Consider a model for $\bomega$ that is polynomial in the elapsed time
\begin{equation}
	\label{eq:omegaLinearModel}
	\bomega\left(t\right) = \sum_{i=1}^Q \bp_i \left(t-t_k\right)^{i-1}
\end{equation}
The model parameters $P_Q\in \mathbb{R}^{3\times Q}$
\begin{equation}
	P_Q = \begin{bmatrix}
		\bp_1 & \bp_2 & \cdots & \bp_Q
	\end{bmatrix}
\end{equation}
can be obtained from $\Delta\btheta$ measurements by solving a linear system of equations. For each measurement, equation by is obtained by replacing $\bomega$ in eq.~\eqref{eq:DefDeltaTheta} with eq.~\eqref{eq:omegaLinearModel}
\begin{subequations}
\begin{align}
	\Delta\btheta_{k+j} &= \int^{t_k  + j \Delta t}_{t_k + \left(j-1\right) \Delta t} \bomega(\tau) d\tau\\
	&=\sum_{i=0}^Q \bp_k \frac{1}{i}\left(\left(j \Delta t\right)^i - \left(j-1\right)^i \Delta t^i\right)
\end{align}
\end{subequations}
From two measurements, an affine model can be formed for $\bomega$. Collecting two measurements in the matrix $\Theta_2\in\mathbb{R}^{3\times2}$
\begin{equation}
	\label{eq:Theta2}
	\Theta_2 = \begin{bmatrix}
		\Delta\btheta_{k-1} & \Delta\btheta_{k}
	\end{bmatrix}
\end{equation}
a linear system of equations is formed
\begin{equation}
	\label{eq:DeltaThetaExpansion:affine}
	\Theta^\top = 
	\begin{bmatrix}
		-\frac{1}{2}\Delta t^2 & \Delta t \\
		\frac{1}{2}\Delta t^2 & \Delta t 
	\end{bmatrix}
	P_2^\top
\end{equation}
Collecting the desired samples $\bomega$ in the matrix $\Omega\in\mathbb{R}^{3\times 3}$
\begin{equation}
	\label{eq:Omega}
	\Omega = \begin{bmatrix}
		\bomega\left(t_k\right) & \bomega\left(t_k + \Delta t/2\right) & \bomega\left(t_k + \Delta t\right)
	\end{bmatrix}
\end{equation}
which can be expressed as the product of a Vandermonde matrix and the model parameters
\begin{equation}
	\label{eq:Q:vander}
	\Omega^\top = V_Q\left(t_k, t_k + \Delta t/2 , t_k + \Delta t \right) P_Q^\top
\end{equation}
As a matter of convention, the polynomial order in the Vandermonde matrix {\it decreases} with column index, such that $V_{i,j} = t_i^{Q-j}$. Solving for $P$ in \eqref{eq:DeltaThetaExpansion:affine} and substituting the result in \eqref{eq:Q:vander}, $\Omega$ is expressed in terms of $\Theta_2$.
\begin{align}
	\label{eq:OmegaQuadratic}
	\Omega^\top &= \begin{bmatrix}
		0 & 1 \\
		\frac{\Delta t}{2} & 1 \\
		\Delta t & 1 \\
	\end{bmatrix}
	\begin{bmatrix}
		-\frac{1}{2}\Delta t^2 & \Delta t \\
		\frac{1}{2}\Delta t^2 & \Delta t
	\end{bmatrix}^{-1}
	\Theta_2^\top\\
	&=\frac{1}{2\Delta t}\begin{bmatrix}
		1 & 1 \\
		0 & 2 \\
		-1 & 3
	\end{bmatrix}
	\Theta_2^\top
\end{align}
Explicitly, 
\begin{subequations}
\label{eq:omega:RK}
\begin{align}
	\label{eq:omega:theta2:k-1}
	\bomega\left(t_k\right) &= \frac{1}{2\Delta t}\left(\Delta \btheta_{k-1} + \Delta \btheta_{k}\right)\\
	\bomega\left(t_k + \frac{\Delta t}{2}\right) &= \frac{1}{\Delta t}\Delta \btheta_{k}\\
	\bomega\left(t_k + \Delta t\right) &= \frac{1}{2\Delta t}\left( 3 \Delta \btheta_{k} - \Delta \btheta_{k-1}\right)
	\label{eq:omega:theta2:k+1}
\end{align}
\end{subequations}
Substituting the expressions from eq.~\eqref{eq:omega:RK} into the second-order RK4 approximation from eq.~\eqref{eq:rk4:omega} reveals
\begin{equation}
	\Delta \bphi_\textrm{RK4}\left(t_k + \Delta t\right) \approx \Delta \btheta_k + \frac{1}{12}\left(\Delta\btheta_{k-1}\times \Delta \btheta_k \right)\label{eq:RK4:Theta2}
\end{equation}
which is precisely the same as the classical result from eq.~\eqref{eq:ConingOneSpeed} (and derived in the Appendix).

The advantage of the proposed framework is that it can easily be extended to higher orders or to more measurements. For example, the previous approach can be extended to three measurements by defining 
\begin{equation}
	\label{eq:Theta3}
	\Theta_3 = \begin{bmatrix}
		\Delta\btheta_{k-1} & \Delta\btheta_{k} & \Delta\btheta_{k+1}
	\end{bmatrix}
\end{equation}
Three measurements can support a quadratic model for $\bomega$, parameterized using 
\begin{equation}
	\label{eq:DeltaThetaExpansion:quadratic}
		P_3^\top = 
	\begin{bmatrix}
		\frac{1}{3} \Delta t^3 & -\frac{1}{2}\Delta t^2 & \Delta t \\
		\frac{1}{3} \Delta t^3 & \frac{1}{2}\Delta t^2 & \Delta t \\
		\frac{7}{3} \Delta t^3 & \frac{3}{2} \Delta t^2 & \Delta t
	\end{bmatrix}^{-1} \Theta_3^\top
\end{equation}
The three samples points for $\bomega$ required by the RK4 algorithms are then given by
\begin{subequations}
\begin{align}
	\Omega^\top &= V_3\left(t_k, t_k + \Delta t/2 , t_k + \Delta t \right) P_3^\top\\
	& = \frac{1}{24\Delta t}\begin{bmatrix}
		8 & 20 & -4 \\
		-1 & 26 & -1 \\
		-4 & 20 & 8
	\end{bmatrix}
	\Theta_3^\top\label{eq:Omega:quad}
\end{align}
\end{subequations}
Replacing $bomega$ in eq.~\eqref{eq:rk4:omega}, using eq.~\eqref{eq:Omega:quad}, results in
\begin{equation}
	\Delta \bphi_\textrm{RK4}\left(t_k + \Delta t\right) \approx \Delta \btheta_k + \frac{1}{288}\left(\Delta\btheta_{k+1}\times \Delta \btheta_{k-1} + 13 \left(\Delta\btheta_{k-1} - \Delta\btheta_{k+1}\right)\times \Delta\btheta_{k} \right)\label{eq:RK4:Theta3}
\end{equation}

The expression \eqref{eq:RK4:Theta3} is an approximation to RK4 integration where the required samples of $\bomega$ were obtained by fitting a curve to three samples of $\Delta \btheta$. This approach could be extended in several ways. For example, additional samples such as $\Delta \btheta_{k-2}$ could yield higher-order curve fit for $\bomega$. Alternatively, curve fitting could be applied in a least-squares sense to mitigate noise or quantization effects in $\Delta \btheta$ measurements.

\section{Results}

Applying sequential rotations using the convention from eq.~\eqref{eq:T:expanded}, 
\begin{equation}
    \bT_{k+1} = \bT( \Delta \bphi_k ) \bT_k
\end{equation}
and so,
\begin{equation}
    \bT( \Delta \bphi_k ) =  \bT_{k+1} \bT_k^{-1}
\end{equation}
Here, $T\left(\Delta\bphi_k\right)\in SO(3)$ is parameterized by $\Delta\bphi$ which is defined in eq.~\eqref{eq:DefDeltaPhi}. The algorithms presented above may be used to compute the parameters $\Delta\bphi$ by solving the ODE of eq.~\eqref{eq:ode_thetadot} from an initial condition $\bphi\left(t_k\right) = 0$. Reconstructions $\Delta\hat{\bphi}$ are obtained from discrete samples of $\bomega$ or $\Delta \btheta$. Algorithm performance is scored against a truth trajectory, $\Delta\bphi^*$. The complete simulation is available on GitHub \cite{HITMAN_2026}.

Instead of defining $\Delta\bphi^*$ analytically, B\'{e}zier curves were used to define $\bomega(t)$ analytically. One benefit is that $\Delta\btheta$, given by eq.~\eqref{eq:DefDeltaTheta}, are available in closed form. Two B\'{e}zier curves were defined for $t\in\left[0,1\right]$: one benign and one challenging. A relatively flat cubic polynomial was chosen for the benign curve. For the challenging curve, control points were selected by randomly generating a $6\times 3$ matrix with standard Gaussian values. For each curve, segments were selected with a random (uniformly sampled) starting location and varying step sizes. Discrete uniformly sampled measurements about the segement of interest were collected in $\Omega$ and $\Theta$ for instantaneous rate and integrated rate measurements, respectively.

Truth trajectories were computed by solving the ODE eq.~\eqref{eq:ode_thetadot} numerically to obtain $\Delta\bphi^*$. Specifically, SciPy \texttt{solve\_ivp} was used with the default RK4(5) variable step-size solver, tolerances reduced to 1e-10, and a max step size of 1e-4. Variable step-size solvers sample $\bomega(t)$ with fine resolution in contrast to the coarse measurements available to algorithms under test.  Alternatively, $\bphi(t)$ could be defined analytically in which case integrated measurements $\Delta\btheta$ must be evaluated numerically. No appreciable changes were observed in contrasting algorithm performance when choosing to define $\bphi(t)$ analytically instead of $\bomega(t)$.

For instantaneous rate measurements, attitude updates were computed by solving the ODE \eqref{eq:f} numerically with algorithm \ref{alg:RK}. Named algorithms \texttt{FwdElr}, \texttt{ExMid}, \texttt{RK3} and \texttt{RK4} employed the Runge-Kutta algorithm \ref{alg:RK} with Tableaux \ref{tbl:fwdeuler}, \ref{tbl:expmid}, \ref{tbl:rk3}, and \ref{tbl:rk4}, respectively. For each tableau, the first column indicates the required sample points of $\bomega$. As a matter of notation, the subscript decoration on $\Omega_n \in \mathbb{R}^{3\times n}$ is used to partition only the first $n$ columns of $\Omega$ defined in eq.~\eqref{eq:Omega}. Notice \texttt{FwdElr} and \texttt{ExMid} operate on $\Omega_1$ and $\Omega_2$, respectively. Both RK3 and RK4 operate on $\Omega_3$. 

For integrated rate measuremments, attitude updates are computed using single-speed algorithms. The \texttt{SingleSpeed} $\Theta_2$ algorithm from eq.~\eqref{eq:RK4:Theta2} is identical to the classical single-speed algorithm from eq.~\eqref{eq:ConingOneSpeed}. It operates on $\Theta_2$ defined in eq.~\eqref{eq:Theta2}. The \texttt{SingleSpeed} $\Theta_3$ uses eq.~\eqref{eq:RK4:Theta3}, operating on $\Theta_3$ defined in eq.~\eqref{eq:Theta3}.

Direct comparison of $\Delta\hat{\bphi}$ are $\Delta\bphi^*$ are problematic due to the non-uniqueness of exponential coordinates. Mapping exponential coordinates back to their group elements avoids artifacts due rotations of $2\pi$. When quantifying algorithm performance, the Frobenius norm of the group elements was employed
\begin{equation}
	\left\| \exp\left( \left(\Delta\hat{\bphi}\right)^{\wedge} \right) - \exp\left( \left(\Delta\bphi^*\right)^{\wedge} \right) \right\|_\textrm{Fro}
\end{equation}

\begin{figure}[h!]
	\centering
	\begin{subfigure}{0.35\textwidth}
		\centering
		\includegraphics[scale=0.8]{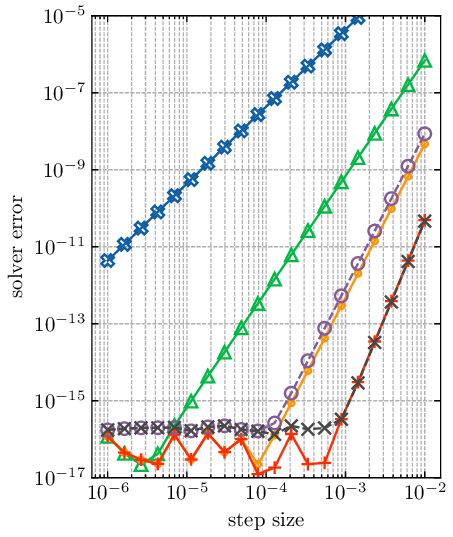} 
		\caption{Benign Trajectory}
		\label{fig:benign}
	\end{subfigure}
	\hfill
	\begin{subfigure}{0.55\textwidth}
		\centering
		\includegraphics[scale=0.8]{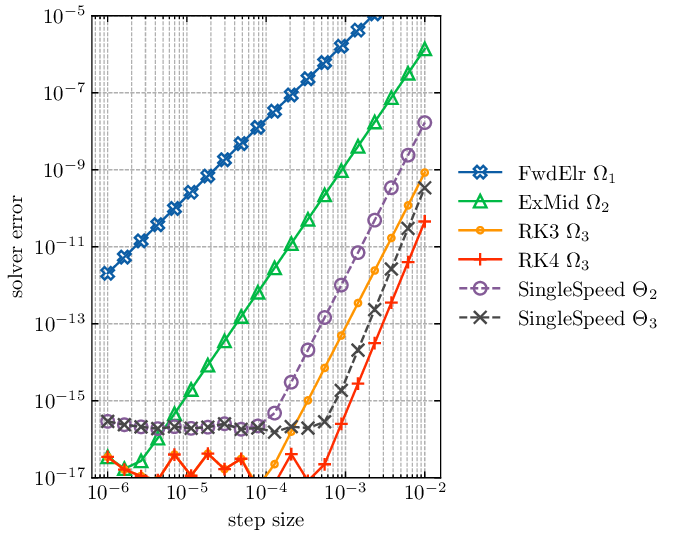} 
		\caption{Challenging Trajectory}
		\label{fig:stress}
	\end{subfigure}
	
	\caption{Algorithm performance on discrete measurements for benign and challenging trajectories. Attitude propagation from instantaneous or integrated rate measurements are indicated by $\Omega_n$ and $\Theta_n$, respectively where the subscript $n$ indicates the number of measurements utilized. \texttt{SingleSpeed} $\Theta_2$ (traditional coning compensation) and $\Theta_3$ use eqs.~\eqref{eq:RK4:Theta2} and \eqref{eq:RK4:Theta3}, respectively.}
	\label{fig:results}
\end{figure}

Results are depicted in Figure \ref{fig:results}. In the legend, $\Omega$ and $\Theta$ are used to distinguish results obtained from instantaneous rates or integrated rate measurements, respectively. The subscripts indicate the number of measurements utilized. Results for \texttt{FwdElr}, \texttt{ExMid}, \texttt{RK3}, and \texttt{RK4} operating on direct samples of $\bomega$ illustrate expected reduction in solver error function with solver order. The improvement in solver error from $\Theta_2$ to $\Theta_3$ illustrates the impact of a quadratic model for $\bomega$. A quadratic model for $\bomega$ is required to recover benefits of a 4th order integrator as indicated by the agreement between RK4 $\bomega$ and SingleSpeed $\Theta_3$ for the benign trajectory. Without the "future" measurement $\Delta\btheta_{k=1}$, the SingleSpeed $\Theta_2$ approximation performs similarly to RK3 $\bomega$. Further, the agreement suggests negligible impact of the approximation eq.~\eqref{eq:Jinv:approx} for the example trajectory and justifies the simplification eq.~\eqref{eq:ConingOneSpeed} even on challenging trajectories.

\section{Conclusion}
This work introduces a new framework for developing coning correction algorithms of varying order and with a varying number of gyro measurements. The framework is developed from Lie theory and approximations of Runge-Kutta methods to solve the requisite differential equations. The classical single-speed coning algorithm popularized by Miller in the early 1980s is found to be a special case of an RK4 integrator. To fully realize the benefits of a 4th order model, however, requires the addition of an additional prior gyro measurement. The methods presented in this work may be straightforwardly extended to higher-order (using the appropriate Butcher tableau) and to more measurements.


\section*{Acknowledgments}
Sandia National Laboratories is a multimission laboratory managed and operated by National Technology \& Engineering Solutions of Sandia, LLC, a wholly owned subsidiary of Honeywell International Inc., for the U.S. Department of Energy’s National Nuclear Security Administration under contract DE-NA0003525.

This paper describes objective technical results and analysis. Any subjective views or opinions that might be expressed in the paper do not necessarily represent the views of the U.S. Department of Energy or the United States Government.

\setcounter{secnumdepth}{-1}
\section{Appendix: Analytic Derivation of Single-Speed Coning Correction}
\label{app:SingleSpeed}

This appendix summarizes a direct derivation for the coning correction in eq.~\eqref{eq:ConingOneSpeed} for a linearly varying angular velocity. Without loss of generality, suppose that $\bomega$ is given by
\begin{equation}
\label{eq:omega_affine}
	\bomega = \bp_2 t + \bp_1
\end{equation}
over the interval $t\in [-\Delta t, \Delta t]$ and where $\{ \bp_i \}_{i=1}^2 \in \RR^3$ are constant vectors that are unknown. Now, \eqref{eq:omega_affine} may be used to analytically compute $\btheta(t)$ from eq.~\eqref{eq:PhiIntegralDef} as
\begin{equation}
	\btheta = \int_{0}^{t} \bomega \, d \tau = \int_{0}^{ t}  \bp_2 \tau + \bp_1 \, d \tau = \frac{1}{2} \bp_2 {t}^2 + \bp_1 { t} + \bc = \frac{1}{2} \bp_2 { t}^2 + \bp_1{ t} 
\end{equation}
where the constant of integration is $\bc = 0$ since the initial condition is $\btheta(0) = \textbf{0}$. It follows, therefore, that
\begin{equation}
	\btheta \times \bomega  = \left( \frac{1}{2} \bp_2 { t}^2 + \bp_1 t \right) \times \left( \bp_2 {t} + \bp_1 \right) = \frac{1}{2} (\bp_2 \times \bp_1) { t}^2 + (\bp_1 \times \bp_2)  { t}^2 = \frac{1}{2} (\bp_1 \times \bp_2)  { t}^2
\end{equation}
which made use of the cross-product identity $\bp_2 \times \bp_1 = - \bp_1 \times \bp_2$. It is now straightforward to analytically compute the coning integral from eq.~\eqref{eq:ConingGR}
\begin{equation}
	\label{eq:LinearOmegaConingIntegralStep1}
	\Delta \bbeta =  \frac{1}{2} \int_{0}^{\Delta t} \btheta \times \bomega \, dt 
		=  \frac{1}{4} (\bp_1 \times \bp_2) \int_{0}^{\Delta t}  t^2 \, dt = \frac{1}{12}  (\bp_1 \times \bp_2) \Delta t^3
\end{equation}

The trouble is that $\bp_1$ and $\bp_2$ are unknown and, therefore, it is not (yet) possible to evaluate eq.~\eqref{eq:LinearOmegaConingIntegralStep1} to find $\Delta \bbeta$. To circumvent this problem, first compute the actual gyro measurement at the current and prior time steps as
\begin{equation}
\label{eq:DeltaThetaExpansion:k}
	\Delta \btheta_k = \btheta(\Delta t) =  \frac{1}{2} \bp_2 \, \Delta t^2 + \bp_1 \, \Delta t
\end{equation}
\begin{equation}
\label{eq:DeltaThetaExpansion:km1}
	\Delta \btheta_{k-1} = -\btheta(-\Delta t) =  -\frac{1}{2} \bp_2 \, \Delta t^2 + \bp_1 \, \Delta t
\end{equation}
from which it is observed that
\begin{align}
	\Delta \btheta_{k-1} \times \Delta \btheta_k & = \left(  -\frac{1}{2} \bp_2 \, \Delta t^2 + \bp_1 \, \Delta t \right) \times \left(  \frac{1}{2} \bp_2 \, \Delta t^2 + \bp_1 \, \Delta t \right) \\
 & = -\frac{1}{2}(\bp_2 \times \bp_1) \Delta t^3 + \frac{1}{2}(\bp_1 \times \bp_2) \Delta t^3  \nonumber \\
& = (\bp_1 \times \bp_2) \Delta t^3 \nonumber
\end{align}
The final relationship is recognized to be precisely the unknown term in the coning integral from eq.~\eqref{eq:LinearOmegaConingIntegralStep1}. Therefore, substituting this result to compute the coning correction effects
\begin{equation}
	\label{eq:SensorIntervalConing}
	\Delta \bbeta_k \approx \frac{1}{12} \Delta \btheta_{k-1} \times \Delta \btheta_k
\end{equation}
which is precisely the single-speed conning correction from eq.~\eqref{eq:ConingOneSpeed}.


\bibliography{imurefs}

\end{document}